\documentclass[letterpaper, 10 pt, conference]{ieeeconf} % Comment this line out if you need a4paper
\IEEEoverridecommandlockouts    
\usepackage[english]{babel}
\usepackage{hyperref}
\usepackage{algorithmic}
\usepackage{balance}
\usepackage{graphicx} % for pdf, bitmapped graphics files
\graphicspath{{ {./images/} }}
\usepackage{epsfig} % for postscript graphics files
\usepackage{svg}
\usepackage{amsmath} % assumes amsmath package installed
\usepackage{amssymb}  % assumes amsmath package installed
\usepackage{bm}
\usepackage{color}
\usepackage{subcaption}
\usepackage{caption}
\usepackage{cite}
\usepackage{rotating}
\usepackage{multicol}
\usepackage{multirow}
\usepackage{textcomp}
\def\BibTeX{{\rm B\kern-.05em{\sc i\kern-.025em b}\kern-.08em
    T\kern-.1667em\lower.7ex\hbox{E}\kern-.125emX}}
\title{\LARGE \bf Generalized Momenta-Based Koopman Formalism for Robust Control of Euler-Lagrangian Systems}

\author{
Rajpal~Singh$^1$,~Aditya~Singh$^1$, ~Chidre~Shravista~Kashyap$^2$,~and~Jishnu~Keshavan$^1$
\thanks{This research is funded by Ministry of Mines Research Grant.}% <-this % stops a space
\thanks{$^1$Department of Mechanical Engineering, Indian Institute of Science, Bengaluru, Karnataka~560012, India 
     ({\tt\footnotesize{email: rajpalsingh@iisc.ac.in, adityasingh2@iisc.ac.in,  kjishnu@iisc.ac.in}}).
     }
\thanks{$^2$Robert Bosch Centre for Cyber-Physical Systems, Indian Institute of Science, Bengaluru, Karnataka~560012, India 
     ({\tt\footnotesize{email:chidres@iisc.ac.in}})
     }
}

\newtheorem{remark}{Remark}
\begin{document}
\maketitle
\begin{abstract}

This paper presents a novel Koopman operator formulation for Euler–Lagrangian dynamics that employs an implicit generalized momentum-based state space representation, which decouples a known linear actuation channel from state-dependent dynamics and makes the system more amenable to linear Koopman modeling. By leveraging this structural separation, the proposed formulation only requires to learn the unactuated dynamics rather than the complete actuation-dependent system, thereby significantly reducing the number of learnable parameters, improving data efficiency, and lowering overall model complexity. In contrast, conventional explicit formulations inherently couple inputs with the state-dependent terms in a nonlinear manner, making them more suitable for bilinear Koopman models, which are more computationally expensive to train and deploy. Notably, the proposed scheme enables the formulation of linear models that achieve superior prediction performance compared to conventional bilinear models while remaining substantially more efficient. To realize this framework, we present two neural network architectures that construct Koopman embeddings from actuated or unactuated data, enabling flexible and efficient modeling across different tasks. Robustness is ensured through the integration of a linear Generalized Extended State Observer (GESO), which explicitly estimates disturbances and compensates for them in real-time. The combined momentum-based Koopman and GESO framework is validated through comprehensive trajectory tracking simulations and experiments on robotic manipulators, demonstrating superior accuracy, robustness, and learning efficiency relative to state-of-the-art alternatives.

\end{abstract}

\section{Introduction}
Control of nonlinear systems has long been a central challenge in robotics. Design of control methodologies that are general, scalable, and computationally efficient is particularly difficult owing to the complexity and high-dimensionality of such systems. Traditional model-based approaches often employ local linearization to make the problem tractable. However, local linearization techniques provide valid approximations only in the vicinity of nominal operating points and fail to represent system behavior precisely over the entire workspace.

In contrast, Koopman operator theory~\cite{koopman1931hamiltonian} offers a globally linear formulation for nonlinear dynamics by leveraging nonlinear observable/basis functions of the state of the system, enabling significantly enhanced prediction accuracy~\cite{korda2018linear}. However, an exact representation of the Koopman operator often requires an infinite dimensional set of observables, which renders it infeasible for practical applications. To address this, several data-driven approaches such as Dynamic Mode Decomposition (DMD)~\cite{dmdtu2013}, DMDc~\cite{dmdcproctor2016}, Extended DMD (EDMD)~\cite{edmdwilliams2014}, and Sparse Identification of Nonlinear Dynamics (SINDy)~\cite{sindybrunton2016} have been proposed to approximate the Koopman operator with a finite set of basis functions. However, selecting a set of observables that is both sufficiently expressive and computationally tractable remains a significant challenge, particularly for complex dynamical systems. As such, recent advances have turned to learning-based methods~\cite{li2017extended, champion2019data} to automate the discovery of an optimal set of basis functions.

% Despite these advances, existing Koopman approaches often attempt to learn the full closed-loop or actuation-dependent dynamics, leading to high model complexity, large parameter counts, and increased data requirements. Moreover, these models are often sensitive to external disturbances and unmodeled dynamics, limiting their robustness in real-world settings.

Euler–Lagrangian systems constitute a core class of nonlinear dynamical systems widely encountered in robotics~{\cite{craig2009introduction}}. While Koopman-based formulations have shown promise, applying them directly to the modeling of the Euler-Lagrangian dynamical systems poses specific challenges. When represented \textit{explicitly in terms of generalized positions and velocities}, the dynamics of Euler–Lagrangian systems is characterized by state-dependent control vector fields. Consequently, this dynamics is naturally more suited for learning bilinear Koopman representations~\cite{goswami2017global}, which can capture nonlinear interactions between state and control terms, rather than purely linear Koopman representations. Hence, many recent works have employed bilinear Koopman embedding to model Euler-Lagrangian systems~\cite{sah2024real, bruder2021advantages}. However, these bilinear models are significantly more complex to train, require larger datasets, and render the downstream control synthesis, especially for optimal control frameworks such as Model Predictive Control (MPC), more complicated and computationally demanding. To address these issues, the authors in~\cite{asada2024control} construct Koopman models at the actuator dynamics level rather than at the joint dynamics level. At the actuator level, the control inputs are naturally decoupled from the state-dependent dynamics, which makes the system dynamics more amenable to learning linear embeddings. However, this design choice requires modeling the influence of actuation indirectly by incorporating the actuator subsystem into the Koopman lifting so that the effects of actuation appear linearly. While this approach preserves the known mechanical structure, it complicates model training and introduces sensitivity to actuator dynamics and parameter uncertainty, which could otherwise be avoided by modeling directly at the joint dynamics level.

Motivated by this challenge, we propose a momentum-based Koopman operator framework tailored specifically for modeling Euler-Lagrangian system dynamics. Instead of using the general explicit state space representation, we employ an \textit{implicit state space representation}, which uses generalized position and generalized momentum as generalized coordinates. This implicit state representation decouples a linear control channel from the remaining state-dependent nonlinear dynamics, making it more amenable to learning linear Koopman models. As such, the linear models obtained from the proposed formulation achieve superior prediction performance relative to bilinear models derived from explicit state representations. Further, the structure of the control matrix can be inferred a priori. As such, only the unactuated/passive part of the dynamics needs to be learned, drastically reducing the number of learnable parameters and improving data efficiency without sacrificing predictive power. We design two neural network architectures capable of learning Koopman embeddings from either actuated or purely unactuated data, enabling flexible modeling under diverse data collection scenarios. Another critical challenge in data-driven Koopman modeling is that the learned models are often sensitive to external disturbances and unmodeled dynamics, which prevents accurate prediction and control, limiting their robustness in real-world settings. As such, previous works have been proposed to improve the robustness of learned Koopman models. In particular,~\cite{lyu2025koopman} and~\cite{chen2024incorporating} employ higher-order Generalized Extended State Observer (GESO)~\cite{han2009pid} in order to deal with external disturbance. Inspired by the previous works, we integrate a linear GESO within our framework to estimate and compensate for external disturbances and unmodeled effects in real time. Unlike the nonlinear higher-order GESOs~\cite{lyu2025koopman, chen2024incorporating} employed in previous studies, linear GESOs can be seamlessly incorporated within the linear controller frameworks without breaking their linear structure, thereby avoiding the additional complications such as saturation filters or stability adjustments. The benefits of the proposed scheme are demonstrated through extensive simulation and experimental studies, including comparison with leading alternative designs. 

The main contributions are summarized as follows:
\begin{itemize}
    \item {We reformulate the Euler-Lagrangian dynamics in an implicit momentum-based representation that decouples control from state-dependent dynamics, enabling the learning of linear Koopman models which exhibit superior prediction performance compared to the more computationally demanding bilinear models learned using conventional explicit representation.} 
    
    \item By exploiting the known structure of the control matrix, our framework needs to learn only the passive dynamics rather than the full actuation-dependent dynamics, thereby substantially reducing the number of learnable parameters, improving data efficiency, and lowering model complexity.
    
    \item We design two neural network architectures that learn Koopman embeddings from actuated or unactuated (passive) data, enabling flexible deployment under diverse data collection scenarios.
    
    \item We integrate a linear Generalized Extended State Observer (GESO) into the Koopman-MPC framework to explicitly estimate and compensate for external disturbances and unmodeled effects in real-time, while preserving the linear structure of downstream controllers.
    
    % \item We validate the proposed approach through extensive simulation and experimental studies to demonstrate the superior tracking accuracy and disturbance rejection compared to state-of-the-art methods.
\end{itemize}

The remainder of the paper is organized as follows. Section~\ref{sec:prelim} presents the preliminaries for Koopman operator theory. Sections~\ref{sec:formulation} and~\ref{sec:neural_nets} present the proposed momentum-based Koopman formulation and neural architectures employed for learning the subsequent Koopman models. Sections~\ref{sec:geso} and~\ref{sec:MPC} detail GESO formulation for disturbance estimation with its subsequent integration within a MPC framework. Section~\ref{sec:results} describes simulation and experimental results. Section~\ref{sec:conclusion} concludes the paper.

\section{Preliminaries}\label{sec:prelim}
This section presents the fundamental concepts of Koopman operator theory. Consider a nonlinear autonomous system given by 
\begin{eqnarray}
    \label{eq:base_non_linear_sytem}
    \boldsymbol{\dot{x}} = \boldsymbol{f}(\boldsymbol{x}),
\end{eqnarray}
where the state vector $\boldsymbol{x} \in \mathbb{X} \subset \mathbb{R}^{n}$ and the dynamics $\boldsymbol{f}:\mathbb{X} \to \mathbb{X}$ is assumed to be Lipschitz continuous over  $\mathbb{X}$. The corresponding discrete-time dynamics can be written as $\boldsymbol{x}_{k+1} = \boldsymbol{F}(\boldsymbol{x}_k)$, where $\boldsymbol{F} : \mathbb{X} \to \mathbb{X}$ denotes the system’s flow map and $\boldsymbol{x}_k$ represents the state at $k^{th}$ time step.

As per Koopman operator theory~\cite{koopman1931hamiltonian}, the evolution of a set of observable functions $\sigma: \mathbb{X} \to \mathbb{C}$, within a Hilbert space, can be described by a linear, infinite-dimensional Koopman operator $\mathcal{K}: \mathbb{C} \to \mathbb{C}$ such that
\begin{eqnarray}
    \mathcal{K} \circ \sigma(\boldsymbol{x}_k) = \sigma(\boldsymbol{F}(\boldsymbol{x}_k)) = \sigma(\boldsymbol{x}_{k+1}), \nonumber
\end{eqnarray}
where $\circ$ denotes functional composition. 

%While the evolution of these observables is linear in Koopman space, the exact Koopman operator is generally infinite-dimensional, necessitating finite-dimensional approximations for practical applications.

The Koopman operator framework can be extended to systems with control inputs. Consider a control-affine system described by
\begin{eqnarray}
    \label{eq:base_affine}
    \boldsymbol{\dot{x}} = \boldsymbol{f}_0(\boldsymbol{x}) + \sum_{i=1}^{m} \boldsymbol{f}_i(\boldsymbol{x}) u_i,
\end{eqnarray}
where $u_i \in \mathbb{R}^{m} \; \forall \; i = 1, 2, ..., m $ denotes the control inputs. $\boldsymbol{f}_0: \mathbb{X} \to \mathbb{X} $ and $\boldsymbol{f}_i : \mathbb{X} \to \mathbb{X} \;\forall\; i = 1, 2, ..., m $ represent the drift and state-dependent control vector fields, respectively. The corresponding Koopman representation takes the form:
\begin{align}
    \label{eq: koopman linear representation}
    &\boldsymbol{z} = [{\phi}_1(\boldsymbol{x}), {\phi}_2(\boldsymbol{x}), \ldots, {\phi}_{N}(\boldsymbol{x})]^{\top}, \nonumber \\
    &\boldsymbol{\dot{z}} = \boldsymbol{A}_c \boldsymbol{z} + \boldsymbol{B}_c\boldsymbol{u}, \quad \boldsymbol{x} = \boldsymbol{C}^x \boldsymbol{z},
\end{align}
where $\phi_i : \mathbb{X} \to \mathbb{C} \; \forall \; i = 1, 2, ..., N $ are the observable functions, and $\boldsymbol{A}_c \in \mathbb{R}^{N \times N}$, $\boldsymbol{B}_c \in \mathbb{R}^{N \times m}$ are the continuous-time Koopman model matrices. The matrix $\boldsymbol{C}^x \in \mathbb{R}^{n \times N}$ maps the lifted state $\boldsymbol{z}$ back to the original state space $\boldsymbol{x}$ . Similarly, the corresponding bilinear formulation is given as
\begin{align}
    \label{eq: koopman bilinear representation}
    \boldsymbol{\dot{z}} = \boldsymbol{A}_c \boldsymbol{z} + \boldsymbol{B}_c (\boldsymbol{z} \otimes \boldsymbol{u}), \quad \boldsymbol{x} = \boldsymbol{C}^x \boldsymbol{z},
\end{align}
where $\boldsymbol{B}_c \in \mathbb{R}^{Nm \times m}$. $\otimes$ represents the Kronecker product. The corresponding discrete-time formulations are:
\begin{align}
    \label{eq:basic_lin}
    &\boldsymbol{z}_{k+1} = \boldsymbol{A} \boldsymbol{z}_k + \boldsymbol{B} \boldsymbol{u}_k, \quad \boldsymbol{x}_{k+1} = \boldsymbol{C}^x \boldsymbol{z}_{k+1}, \\
    \label{eq:basic_bilin}
    &\boldsymbol{z}_{k+1} = \boldsymbol{A} \boldsymbol{z}_k + \boldsymbol{B} (\boldsymbol{z}_k\otimes\boldsymbol{u}_k), \quad \boldsymbol{x}_{k+1} = \boldsymbol{C}^x \boldsymbol{z}_{k+1}.
\end{align}
A more comprehensive treatment of the Koopman operators for control-affine systems can be found in~\cite{goswami2017global, singh2024adaptive}.

\section{Generalized Momenta-based Koopman Formulation} \label{sec:formulation}
This section presents a momentum-based formulation of the Koopman operator for Euler-Lagrangian dynamical systems. The general Euler-Lagrangian dynamics is given as
\begin{align}
    \label{eq:man_dyn}
    \boldsymbol{M}(\boldsymbol{q})\boldsymbol{\ddot{q}} + \boldsymbol{C}(\boldsymbol{q}, \,\boldsymbol{\dot{q}})\boldsymbol{\dot{q}} + \boldsymbol{G}(\boldsymbol{q}) = \boldsymbol{\tau}, 
\end{align}
where $\boldsymbol{q} \in \mathbb{R}^{l}$ and $\boldsymbol{\tau} \in \mathbb{R}^{m}$ refer to the position coordinates and control inputs, respectively. $\boldsymbol{M}(\boldsymbol{q}) \in \mathbb{R}^{l \times l}$, $\boldsymbol{C}(\boldsymbol{q}, \,\boldsymbol{\dot{q}})  \in \mathbb{R}^{l \times l} $ and $\boldsymbol{G}(\boldsymbol{q})\in \mathbb{R}^{l}$ represent the mass matrix, the Coriolis terms and the gravity term, respectively. 
Traditionally, position ($\boldsymbol{q}$)  and velocity ($\boldsymbol{\dot{q}}$) coordinates are chosen as generalized coordinates to explicitly represent the state space for the Euler-Lagrangian system. The evolution of the generalized velocities is given as
\begin{align}
    \boldsymbol{\ddot{q}}  =\boldsymbol{M}(\boldsymbol{q})^{-1}\left( \boldsymbol{\tau}   - \boldsymbol{C}(\boldsymbol{q}, \,\boldsymbol{\dot{q}})\boldsymbol{\dot{q}} - \boldsymbol{G}(\boldsymbol{q}) \right) \nonumber.
\end{align}
Due to their interaction with the mass matrix, control vectors become state-dependent. Consequently, this explicit representation lends itself more naturally to bilinear formulation~(\ref{eq: koopman bilinear representation})~\cite{goswami2017global}. Although expressive, bilinear Koopman models are computationally more expensive, require larger datasets, and often preclude integration with linear MPC frameworks~\cite{bruder2021advantages}.

Instead, we employ an implicit momentum-based formulation, where the state space vector is defined as:
\begin{align}
    \label{eq:mom_state_space}
    \boldsymbol{x} = \begin{bmatrix}
    \boldsymbol{q} \\
    \boldsymbol{M}(\boldsymbol{q})\boldsymbol{\dot{q}}
    \end{bmatrix}
    = 
    \begin{bmatrix}
    \boldsymbol{x}_u \\
    \boldsymbol{x}_a
    \end{bmatrix} \in \mathbb{R}^{n}
\end{align}

where $\boldsymbol{x}_u$ and $\boldsymbol{x}_a$ represent the unactuated and the actuated part of the dynamics, respectively. The evolution of the  proposed state space vector is given as
\begin{align}
    \label{eq:mom_state_space_evol}
    \boldsymbol{\dot{x}} 
     & =  \begin{bmatrix}
    \boldsymbol{M}^{-1}\boldsymbol{x}_a\\
    (\boldsymbol{\dot{M}}(\boldsymbol{x}_u) {-} \boldsymbol{C}(\boldsymbol{x_u}, \boldsymbol{M}^{-1}\boldsymbol{x}_a))\boldsymbol{M}^{-1}\boldsymbol{x}_a {-} 
    \boldsymbol{G}(\boldsymbol{x_u}) {+} \boldsymbol{\tau}
    \end{bmatrix} \nonumber \\
    & =  \begin{bmatrix}
    \boldsymbol{f}_u(\boldsymbol{x}) \\
    \boldsymbol{f}_a(\boldsymbol{x}) + \boldsymbol{\tau}
    \end{bmatrix} = \boldsymbol{f}(\boldsymbol{x}) + \boldsymbol{\tilde{u}}, 
\end{align}
where $\boldsymbol{f}(\boldsymbol{x}) = [ \boldsymbol{f}_u(\boldsymbol{x}), \boldsymbol{f}_a(\boldsymbol{x})]^\top$  and  $\boldsymbol{\tilde{u}} = [\boldsymbol{0}_{l}, \boldsymbol{u}]^\top$ where $\boldsymbol{0}_{l} \in \mathbb{R}^{l}$ represents the zero vector and $\boldsymbol{u} = \boldsymbol{\tau}$. Hence, the proposed choice of state space completely decouples the linear control inputs from the state-dependent dynamics, making resultant dynamics more amenable for the formulation of linear Koopman models~(\ref{eq: koopman linear representation}). 

Having separated a linear control channel, we can apply the Control Coherent Koopman modeling formulation~\cite{asada2024control} to construct a linear Koopman model by writing the passive/unactuated part of the dynamics as:
\begin{align}
    \label{eq:passive_dyn}
    \boldsymbol{\dot{x}}_p = \boldsymbol{\dot{x}} -\boldsymbol{\tilde{u}},
\end{align}

Consider the Koopman lifted state to be $\boldsymbol{z} = [\boldsymbol{x}^\top,   \boldsymbol{\phi}(\boldsymbol{x})^\top]^\top \in \mathbb{R}^{N}$,  where $\boldsymbol{\phi}(\cdot): \mathbb{R}^{n} \to \mathbb{R}^{P}$ represents the observable functions. Then, as per~\cite{asada2024control}, we can construct a Koopman matrix to relate the passive and the actuated dynamics as
\begin{align}
    \label{eq:Koop_model_pass}
    \begin{bmatrix}
     \boldsymbol{\dot{x}}_p  \\
    \boldsymbol{\dot{\phi}}(\boldsymbol{x})
     \end{bmatrix} = 
     \begin{bmatrix} \boldsymbol{A}_{c,xx} & {A}_{c,x\phi} \\
                     \boldsymbol{A}_{c,x\phi} & {A}_{c,\phi\phi},       
     \end{bmatrix}
     \begin{bmatrix}
     \boldsymbol{x} \\
    \boldsymbol{{\phi}}(\boldsymbol{x})
     \end{bmatrix}
\end{align}

where the Koopman matrix is expressed as block related to the evolution of state variables ($\boldsymbol{x}$) and the observable functions ($\boldsymbol{\phi}(\boldsymbol{x})$). Substituting~(\ref{eq:passive_dyn}), we get
\begin{align}
    \label{eq:pass_koop_dyn}
    &{\boldsymbol{\dot{z}}} = \boldsymbol{A}_c\boldsymbol{z} + \boldsymbol{B}_c\boldsymbol{u}, \; \boldsymbol{x} = \boldsymbol{C}^x \boldsymbol{z},     
\end{align}

where $\boldsymbol{A}_c = \begin{bmatrix} \boldsymbol{A}_{c,xx} & {A}_{c,x\phi} \\
                     \boldsymbol{A}_{c,x\phi} & {A}_{c,\phi\phi}       
     \end{bmatrix}$, $\boldsymbol{B}_c = \begin{bmatrix}
           \boldsymbol{0}_{l\times m}\\
           \boldsymbol{I}_{l\times m}\\
           \boldsymbol{0}_{P\times m}
\end{bmatrix}$, $ \boldsymbol{C}^x = \begin{bmatrix}
           \boldsymbol{I}_{n\times n} &\boldsymbol{0}_{n\times P}
\end{bmatrix}$. $\boldsymbol{I}_{n\times n}$ represents a $n \times n$ identity matrix. The corresponding discrete time Koopman model is 
 \begin{align}
    \label{eq:pass_koop_dyn_disc}
    &\boldsymbol{z}_{k+1} = \boldsymbol{A}\boldsymbol{z}_k + \boldsymbol{B}\boldsymbol{u}_k, \; \boldsymbol{x}_k = \boldsymbol{C}^x \boldsymbol{z}_k.   
\end{align}
It is to be noted that for Control Coherent Modeling Formulation in~\cite{asada2024control}, Koopman models have to be constructed at the actuator dynamics level as a consequence of explicit state space representation. This indirect approach becomes necessary to capture the influence of actuation in a linear manner on system dynamics, as required by the assumptions to construct the formulation in~(\ref{eq:Koop_model_pass}). By contrast, the implicit state space representation naturally resolves this issue and enables the direct application of linear Koopman operator modeling at the joint dynamics level.

\begin{remark}
     In the proposed formulation~(\ref{eq:pass_koop_dyn_disc}), the input matrix $\boldsymbol{B}$ is known a priori. Hence, only the Koopman operator matrix $\boldsymbol{A}$ governing the passive dynamics needs to be learned. This leads to significant improvement in learning efficiency, as the number of learnable parameters is reduced by $Nm$ and $Nm^2$ compared to linear and bilinear Koopman formulations based on explicit state representations, respectively.
\end{remark}

\begin{figure*}[h] 
\centering
    \includegraphics[width=\linewidth]{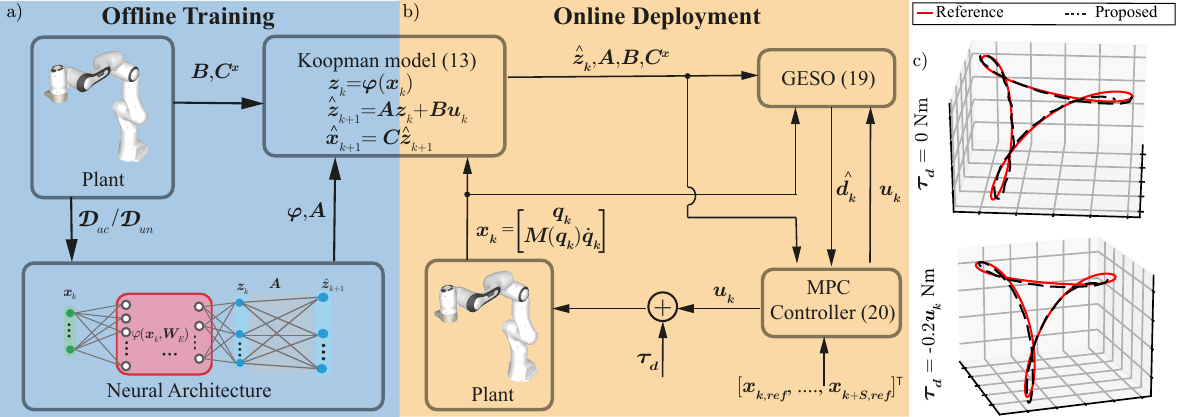}
    \caption{{Framework of the proposed momentum-based Koopman-MPC-GESO architecture depicting a) model training, b) control framework, and c) trajectory tracking results.}
    \label{fig:flow_diag}}
\end{figure*}

\begin{remark} In the momentum-based Koopman model~(\ref{eq:pass_koop_dyn_disc}), the control input $\boldsymbol{u}$ directly influences the momentum states $\boldsymbol{x}_a$ within the same timestep, while its effect on the remaining lifted states $\boldsymbol{z}$ propagates only in the subsequent timestep. This inherent one-step transmission delay is consistent with physical causality analyses~\cite{asada2024control,dean2006system}, where actuation first impacts the actuated subsystem and subsequently propagates through the coupled dynamics.
\end{remark}

% \begin{figure*}[h] 
% \centering
%     \includegraphics[width=\linewidth]{images/passive_koopman_framework.pdf}
%     \caption{{Neural architecture for learning Momentum-based Koopman Operator.}
%     \label{fig:NN}}.
% \end{figure*}

\section{Neural Architectures for learning  Passive Dynamics}\label{sec:neural_nets}

This section describes the neural network architectures developed to learn the passive dynamics of Euler-Lagrangian systems. We introduce two distinct architectures. The first is trained using actuated data, in which control inputs actively influence the evolution of the states; the second is trained on unactuated data, where the system evolves without external actuation. The datasets used for training are defined as follows:
\begin{align}
    \label{eq:dataset}
    \boldsymbol{\mathcal{D}}_{ac} = (\boldsymbol{X}_i, \boldsymbol{Y}_i, \boldsymbol{U}_i)_{i=1}^{p}, \; 
    \boldsymbol{\mathcal{D}}_{un} = (\boldsymbol{X}_i, \boldsymbol{Y}_i)_{i=1}^{p},
\end{align}
where $\boldsymbol{X}_i = [\boldsymbol{x}^{i}_1, \boldsymbol{x}^{i}_2, \ldots, \boldsymbol{x}^{i}_{w-1}]$ and $\boldsymbol{Y}_i = [\boldsymbol{x}^{i}_2, \boldsymbol{x}^{i}_3, \ldots, \boldsymbol{x}^{i}_{w}]$ are the snapshot matrices capturing consecutive state observations, and $\boldsymbol{U}_i = [\boldsymbol{u^{i}}_1, \boldsymbol{u^{i}}_2, \ldots, \boldsymbol{u^{i}}_{w-1}]$ denotes the corresponding control inputs for the actuated dataset. $p$ and $w$ represent the total number of trajectories in the dataset and the number of snapshots per trajectory, respectively.

Both networks share a common loss structure designed to ensure accurate lifting and prediction within the Koopman space. The constituent components of the loss are defined as follows:
\begin{eqnarray}
    \label{eq:offline_nn_loss}
     L_{pred} = \| \boldsymbol{x}_{k+1} {-} \boldsymbol{C}^x \boldsymbol{\hat{z}}_{k+1} \|_2, \;
    L_{lift} =\| \boldsymbol{z}_{k+1} {-} \boldsymbol{\hat{z}}_{k+1} \|_2, 
\end{eqnarray}
where $\boldsymbol{z}_k$ and $\boldsymbol{\hat{z}}_{k}$ denote the ground truth and predicted values of the lifted state. The difference between the two networks lies in how $\boldsymbol{\hat{z}}_{k+1}$ is computed:
\begin{itemize}
    \item For the \textbf{unactuated dataset}, where no control input is applied, $
        \boldsymbol{\hat{z}}_{k+1} = \boldsymbol{A}\boldsymbol{z}_k$.
    \item For the \textbf{actuated dataset}, the evolution also depends on the control input, so $\boldsymbol{\hat{z}}_{k+1} = \boldsymbol{A}\boldsymbol{z}_k + \boldsymbol{B}\boldsymbol{u}_k$.
    Note that, in this case, $\boldsymbol{B}$ is not a learnable parameter but a known matrix.
\end{itemize}
The total training loss is given as
\begin{align}
    L =  \alpha_1 L_{pred} + \alpha_2 L_{lift} + \gamma_1 \|\boldsymbol{W}\|_1 + \gamma_2 \|\boldsymbol{W}\|_2,
\end{align}
where $\alpha_1$ and $\alpha_2$ are weighting coefficients for the prediction and lifting consistency losses, respectively. $\|\boldsymbol{W}\|_1$ and $\|\boldsymbol{W}\|_2$ represent the $L_1$ and $L_2$ norms of the network's trainable parameters, included to encourage sparsity and limit parameter magnitude. Hyperparameters $\gamma_1$ and $\gamma_2$ control the strength of the regularization terms. 

\begin{remark} Datasets collected under unactuated conditions may not achieve the same learning performance as actuated datasets, primarily due to reduced variability and limited excitation of the system dynamics. In particular, the absence of control inputs constrains the richness of the observed trajectories, which can limit the learning performance. Nonetheless, learning from unactuated data remains valuable when actuated data collection is costly, unsafe, or otherwise impractical.
\end{remark}
\begin{figure*}[h!]
     % \begin{subfigure}[b]{0.48\textwidth}
     %     \centering
     %     \includegraphics[width=\textwidth]{images/validation_data_hist_comp.pdf}
     %     \caption{}
     %     \label{fig:valid_comp}
     % \end{subfigure}
     \begin{subfigure}[b]{0.48\textwidth}
         \centering
         \includegraphics[width=\textwidth]{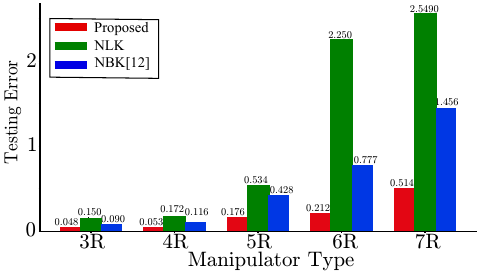}
         \caption{}
         \label{fig:test_comp}
     \end{subfigure}
     % \begin{subfigure}[b]{0.48\textwidth}
     %     \centering
     %     \includegraphics[width=\textwidth]{images/Validation_data_with_data.pdf}
     %     \caption{}
     %     \label{fig:val_data}
     % \end{subfigure}
     \begin{subfigure}[b]{0.48\textwidth}
         \centering
         \includegraphics[width=\textwidth]{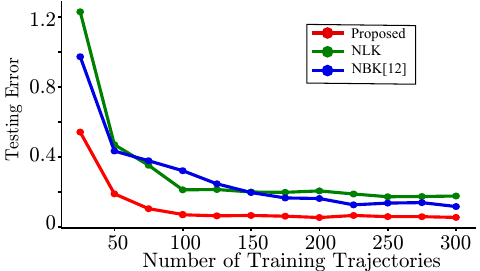}
         \caption{}
         \label{fig:test_data}
     \end{subfigure}
      \caption{{Predictive performance comparison for the proposed scheme with Nominal Linear Koopman (NLK) and Nominal Bilinear Koopman (NBK)~\cite{bruder2021advantages}.  Testing errors across a) different manipulators, and b) different sizes of training datasets for a 4R manipulator.}}
        \label{fig:pred_perf_com}
\end{figure*}
\section{Linear Generalized Extended State Observer}\label{sec:geso}
In practical scenarios, robotic systems are invariably subjected to external disturbances and unmodeled dynamics, possibly arising from the finite-dimensional approximation of the Koopman operator, that can degrade control performance. To enhance robustness against such uncertainties, we incorporate a Generalized Extended State Observer (GESO)~\cite{han2009pid} into the control framework. GESO enables real-time estimation of disturbances, which can then be actively compensated through the control input channel, thereby improving tracking accuracy and robustness.

To this end, consider the Euler-Lagrangain dynamics~(\ref{eq:man_dyn}) in presence of external disturbance, $\boldsymbol{\tau}_d$:
\begin{align}
    \label{eq:man_dyn_disr}
    \boldsymbol{M}(\boldsymbol{q})\boldsymbol{\ddot{q}} + \boldsymbol{C}(\boldsymbol{q}, \,\boldsymbol{\dot{q}})\boldsymbol{\dot{q}} + \boldsymbol{G}(\boldsymbol{q}) = \boldsymbol{\tau} + \boldsymbol{\tau}_d, 
\end{align}

Under the effect of the external disturbance and unmodeled dynamics, the Koopman dynamics~(\ref{eq:pass_koop_dyn}) can now be written as:
\begin{align}
    \label{eq:geso}
    &\boldsymbol{\dot{x}} = \boldsymbol{C}^x(\boldsymbol{A}_c\boldsymbol{z} + \boldsymbol{B}_c\boldsymbol{u}) +  \boldsymbol{{d}},
     \end{align}
where $\boldsymbol{{d}} = \boldsymbol{C}^x\boldsymbol{B}_c \boldsymbol{\tau}_d + \boldsymbol{\Delta}$ is the lumped disturbance term which represents the combined effect of model uncertainty $\boldsymbol{\Delta}$ and external disturbance $\boldsymbol{\tau}_d$. Subsequently, the GESO dynamics can be formulated as
 \begin{flalign}
    \label{eq:geso}
    &\boldsymbol{\dot{\hat{x}}} = \boldsymbol{C}^x(\boldsymbol{A}_c\boldsymbol{z} + \boldsymbol{B}_c\boldsymbol{u}) + {k_1}(\boldsymbol{{x}} - \boldsymbol{\hat{x}}) + \boldsymbol{\hat{d}}, \nonumber \\
    &\boldsymbol{\dot{\hat{d}}} = {k_2}(\boldsymbol{{x}} - \boldsymbol{\hat{x}}),
 \end{flalign}
 where $\boldsymbol{{x}}$ and $\boldsymbol{\hat{d}}$ represent the actual state of the system and the lumped disturbance estimated by linear GESO, respectively. Once the lumped disturbance is estimated, the updated model can be fed to the controller for disturbance compensation.
 %\csk{Additionally, the convergence guarantee of GESO is studied in \cite{}}.

\section{Model Predictive Control}
\label{sec:MPC}
The proposed Koopman architecture is paired with a linear MPC. Linear MPC with a prediction (and control) horizon $S$ is defined as an optimization problem of the form  

\begin{align}
    \label{eq:linear_MPC}
    &\min_{Z,U} \sum_{k=0}^{S}\begin{bmatrix}
     \boldsymbol{C}^x\boldsymbol{z}_k - \boldsymbol{x}_{ref,k}\\
     \boldsymbol{u}_k
    \end{bmatrix}^T \boldsymbol{R} \begin{bmatrix}
     \boldsymbol{C}^x\boldsymbol{z}_k - \boldsymbol{x}_{ref,k}\\
     \boldsymbol{u}_k
    \end{bmatrix} \\
    &\text{s.t} \,\, \boldsymbol{z}_{k{+}1} {=} \boldsymbol{A}\boldsymbol{z}_k {+} \boldsymbol{B}\boldsymbol{u}_k + \boldsymbol{C}^{x \dagger}\boldsymbol{\hat{d}}_kdt, \,\boldsymbol{z}_0 = \boldsymbol{\varphi}(\boldsymbol{x}_0),\nonumber\\
    &\boldsymbol{x}^{-} \leq \boldsymbol{C}^{x}\boldsymbol{z}_k \leq \boldsymbol{x}^{+}, \, \boldsymbol{u}^{-} \leq \boldsymbol{u}_k \leq \boldsymbol{u}^{+} ,\, k = 0,\, ...,\, S-1, \nonumber 
\end{align}
where $\boldsymbol{x}_{ref,k}$ refers to the reference state at $k^{th}$ timestep. $\boldsymbol{R}$ represents the penalty matrix, while $[\boldsymbol{x}^{-},\,\boldsymbol{x}^{+}]$ and $[\boldsymbol{u}^{-},\,\boldsymbol{u}^{+}]$ represent the state and input constraints, respectively. $\dagger$ represents the pseudoinverse. The full proposed momentum-based Koopman-MPC-GESO architecture is shown in Fig.~\ref{fig:flow_diag}.

\begin{figure*}[h] 
\centering
    \includegraphics[width=\linewidth]{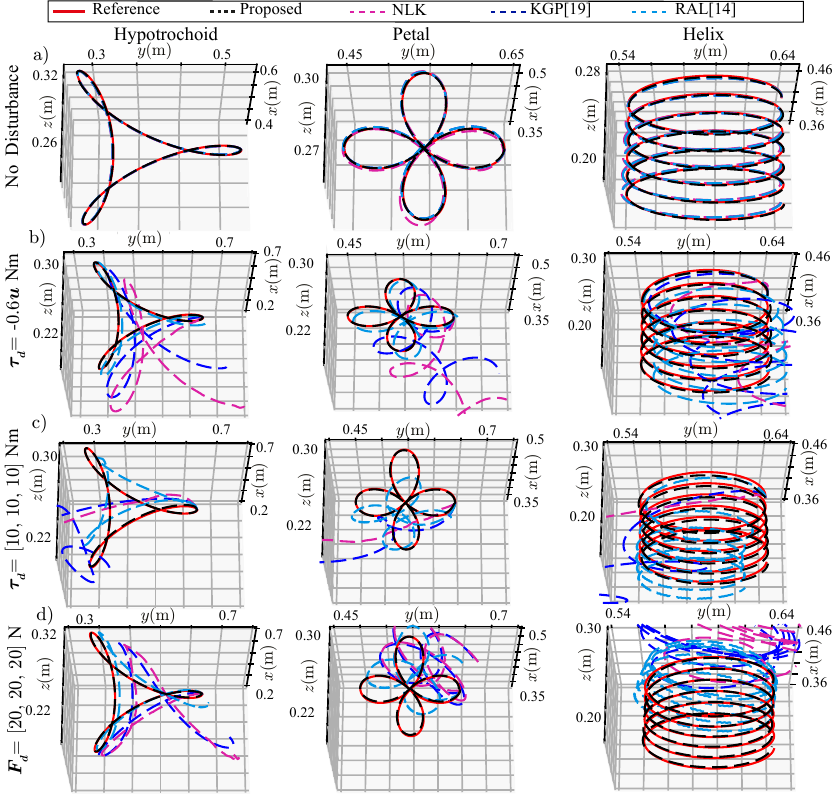}
    \caption{Tracking performance comparisons for a 3R serial manipulator with a) no disturbance, b) actuator fault ($\boldsymbol{\tau_d}= -0.6\boldsymbol{u} \; \text{Nm}$), c) constant disturbance torque ($\boldsymbol{\tau_d} = [10, 10, 10] \;  \text{Nm})$, and d) constant end-effector load(${\boldsymbol{F_d} = [20, 20, 20]\;\text{N}}$). }
    \label{fig:tracking_perf_comp}
\end{figure*}
 \begin{remark}
 The formulation of GESO  as a linear observer allows it to be directly embedded within the MPC architecture while preserving the controller linearity. This integration allows disturbance estimation and compensation to be seamlessly integrated within the optimization solver, such that the state and input constraints are now enforced consistently within the MPC framework. By contrast, the method in~\cite{lyu2025koopman} employs a nonlinear GESO, which cannot be incorporated into the linear MPC formulation without compromising the tracking performance. In particular, the observer is placed outside the control loop, forcing disturbances to be compensated in an ad-hoc manner. This separation not only undermines the controller’s ability to enforce constraints systematically but also requires incorporating additional mechanisms, such as external saturation filters, to maintain safety, thus increasing complexity and potentially reducing overall robustness.
 \end{remark}

\section{Results}\label{sec:results}
This section presents a comprehensive evaluation of the proposed architecture through simulation and experimental studies on serial manipulators. The assessment considers both open-loop prediction accuracy and closed-loop control performance, thereby providing a rigorous demonstration of the effectiveness and robustness of the proposed framework\footnote{code : \href{https://github.com/Rajpal9/Robust_Gen_Mom_Koopman.git}{https://github.com/Rajpal9/Robust\_Gen\_Mom\_Koopman.git}}. 

\subsection{Prediction Performance Comparison}\label{sec:pred_per}
This subsection evaluates the open-loop prediction performance of the proposed scheme. Comparisons are conducted against two baseline approaches, the Nominal Linear Koopman (NLK) model and the Nominal Bilinear Koopman (NBK)~\cite{bruder2021advantages} model. Both baselines employ explicit state space representations, which is in contrast with the current study that relies on an implicit state representation. The datasets are collected at a simulation frequency of $100\; \text{Hz}$ with $1000$ snapshots in each trajectory.

Testing errors are used as comparison metrics. The testing errors are standardized to ensure fair comparison between the proposed scheme ($\boldsymbol{x} = [\boldsymbol{q}^\top, (\boldsymbol{M\dot{q}})^\top]^{\top}$) and the baseline nominal scheme ($\boldsymbol{x} = [\boldsymbol{q}^\top, \boldsymbol{\dot{q}}^\top]^{\top}$), which employ different state space coordinates. The corresponding performance comparison across different ranges of manipulators and sizes of datasets is shown in Fig.~{\ref{fig:pred_perf_com}}. The proposed scheme outperforms both NLK and, more importantly, NBK~\cite{bruder2021advantages} across all the evaluations. This result is particularly noteworthy because bilinear formulations are typically regarded as better suited for Euler-Lagrangian systems due to the intrinsic coupling between states and inputs. Nevertheless, the momentum-based formulation achieves superior performance by structurally separating the linear input channel from the passive nonlinear dynamics, thereby making the system more amenable to linear Koopman modeling. This structural advantage not only reduces modeling complexity but also enables higher prediction accuracy than bilinear approaches, reconciling the dual objectives of modeling accuracy and computational efficiency in a way that bilinear approaches cannot. Further, it can be observed (Fig.~\ref{fig:pred_perf_com}b) that the linear models only require 100 trajectories to achieve the optimal performance, with the proposed scheme consistently outperforming the baselines across all dataset sizes. While NBK~\cite{bruder2021advantages} requires a much larger dataset with about 250 trajectories to achieve convergence in terms of prediction capabilities while still offering considerably inferior performance.  

Hence, the proposed formulation enables learning of linear Koopman models which offer superior performance as compared to bilinear Koopman models based on explicit state representation while offering higher data efficiency. 
% Figs.~{\ref{fig:pred_perf_com}}c shows a comparison of training times which seems to be similar for all the three schemes, with the proposed scheme doing marginally better than the baselines. Figs.~{\ref{fig:pred_perf_com}}d shows the reduction in learnable parameters as compared to NLK which is equal to $Nm$ ranging from 90 in 3R to 770 in 7R. In case of bilinear models, it would be $Nm^2$ less parameters.

% \begin{figure}[ht!]
%     \includegraphics[0.48*width=\textwidth]{images/state_diff.pdf}  
%     \caption{Reduction in learnable parameters achieved by the proposed momentum-based formulation compared to the traditional Koopman approach.} 
%     \label{fig:state_diff}
% \end{figure}
\subsection{Closed Loop Control Performance Comparison}\label{sec:close_per}

\begin{figure*}[h] 
\centering
    \includegraphics[width=\linewidth]{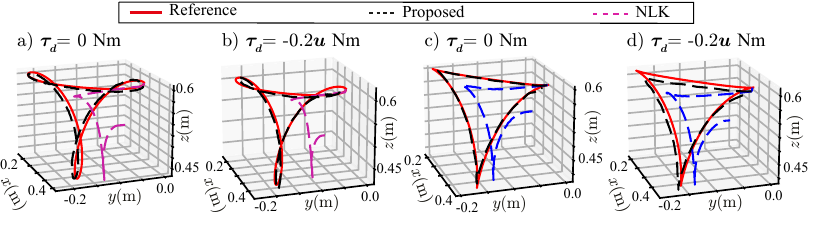}
    \caption{{Experimental results for trajectory tracking with the FR 3 manipulator, a,c) without, and b,d) with external disturbances.}
    \label{fig:franka}}
\end{figure*}

% \begin{figure*}[ht!]
%     \includegraphics[width=\textwidth]{images/combined_err_comp.pdf}  
%     \caption{Performance comparison across different number of training trajectories. a) Open loop testing error. b) Closed loop position tracking error. c)Closed loop full state tracking error.} 
%     \label{fig:val_loss_vs_dim}
% \end{figure*}
This subsection evaluates the closed-loop tracking performance of the proposed scheme through extensive simulations on a serial manipulator. Specifically, we consider a 3R serial manipulator operating in a 3D workspace, with physical parameters defined as $m_i = 0.6 \;\text{kg}, \; l_i = 0.33\;\text{m}, \; I_i = \text{diag}\left[0, \tfrac{m_{i}l_{i}^{2}}{12}, \tfrac{m_{i}l_{i}^{2}}{12}\right]\text{kgm$^2$}$ for $i = 1,2,3$, where $m_i$, $l_i$, and $I_i$ denote the mass, length, and inertia of the $i^{\text{th}}$ link, respectively. The simulations are run at a frequency of 100 Hz. The parameters for GESO are $k_1 = 40$ and $k_2 = 800$. 
To demonstrate the efficacy of the proposed architecture, we conduct comparison studies against state-of-the-art alternatives, including the NLK formulation, Koopman models enhanced with Gaussian Processes (KGP)\cite{lu2024vector}, and the Robust Active Learning (RAL) algorithm\cite{lyu2025koopman}, which constructs an offline model via active learning and augments online control using higher-order GESOs. For fairness, MPC is employed as the base controller across NLK, KGP~\cite{lu2024vector}, and RAL~\cite{lyu2025koopman}. In the previous subsection, we have already demonstrated superior prediction performance over the bilinear models. Further, the computation burden of bilinear models paired with MPC  leading to their infeasibility for real-time applications has already been studied extensively~\cite{sah2024real, singh2024adaptive}. Hence, we restrict ourselves to comparisons with linear models.

Initially, tracking performance evaluation is conducted for a scenario where the controller is tasked with following various reference trajectories in joint space in the absence of any external disturbances. The corresponding results are shown in Fig.~{\ref{fig:tracking_perf_comp}}a and Table~\ref{tab:losses}, which demonstrate the ability of the proposed architecture to accurately trace the reference path. It is noticed that the proposed momentum-based Koopman architecture provides better tracking performance compared to the nominal Koopman model. The superior performance relative to the nominal Koopman model can be attributable to its more precise model. Further, the proposed scheme gives similar or slightly better performance as compared to KGP~\cite{lu2024vector}  and RAL schemes~\cite{lyu2025koopman}.

Subsequently, we consider the tracking evaluation for trajectory tracking under the influence of various types of disturbance. In particular, we consider three types of disturbances namely, actuator faults
($\boldsymbol{d}_1$) with $\boldsymbol{\tau}_d = -0.6\boldsymbol{u}~Nm$, constant torque disturbance  ($\boldsymbol{d}_2$) with ${\boldsymbol{\tau}_d = [10, 10, 10]~Nm}$, and constant end effector load 
($\boldsymbol{d}_3$)  
of $ {\boldsymbol{F_d} = [20, 20, 20]~N}$. The corresponding results are shown in  Figs.~\ref{fig:tracking_perf_comp}b -~\ref{fig:tracking_perf_comp}d and Table~\ref{tab:losses}. As can be seen, even under such extreme external disturbances, the proposed framework is able to track the desired trajectory effectively. This robustness is achieved by the incorporated GESO, which compensates for disturbances in real-time. In contrast, the baseline NLK architecture fails to mitigate the effects of these disturbances, resulting in degraded tracking performance and significant deviation from the desired path. Further, the proposed scheme also significantly outperforms  KGP~\cite{lu2024vector}  and RAL~\cite{lyu2025koopman} schemes, both of which are equipped with capabilities, Gaussian processes and higher-order GESOs respectively, to deal with disturbances. The relatively poor performance of RAL~\cite{lyu2025koopman} can be attributed to the limited prediction accuracy of the offline Koopman model obtained through active learning, which is inferior to the proposed scheme. Furthermore, since RAL~\cite{lyu2025koopman} employs higher-order GESOs, these observers cannot be directly embedded within an MPC framework. Instead, they must be appended to the MPC output, necessitating external saturation filters for safety, which acts as an additional hindrance to the overall effectiveness of RAL~\cite{lyu2025koopman}. In addition, when paired with a high-gain base controller, higher-order GESOs in RAL~\cite{lyu2025koopman} induce control chattering which introduces discontinuous corrective actions that further degrade closed-loop performance. 
\begin{table}[ht!]
\small\centering
\caption{\label{tab:losses} RMSE errors (m) for a variety of disturbances ($\boldsymbol{d_0}$ : No Disturbance, $\boldsymbol{d_1} : \boldsymbol{\tau_d} = -0.6\boldsymbol{u}~Nm$, $\boldsymbol{d_2} : {\boldsymbol{\tau_d} = [10, 10, 10]~Nm}$, $\boldsymbol{d_3}:{\boldsymbol{F_d} = [20, 20, 20]~N}$.)}
    \begin{tabular}{|p{1.6 cm}|p{1.2 cm}|p{1.2 cm}|p{1.2 cm}|p{1.2 cm}| } 
     \hline
      {Disturbance}& {Proposed} & {NLK} & {KGP~\cite{lu2024vector}}&{RAL~\cite{lyu2025koopman}}  \\ 
     \hline
     \multicolumn{5}{|c|}{Hypotrochoid} \\ \hline
     $\boldsymbol{d_0}$ & \textbf{0.0028} & {0.0138} & {0.0040} & {0.0030}\\ \hline
    $\boldsymbol{d_1}$ & \textbf{0.0029} & {0.3507}& {0.1133} & {0.0117} \\ \hline
    $\boldsymbol{d_2}$ &   \textbf{0.0045} & {1.3218} & {0.5736} & {0.0739}\\  \hline
    $\boldsymbol{d_3}$ & \textbf{0.0034} & {0.4573}& {0.1863} & {0.0764} \\ \hline
    \multicolumn{5}{|c|}{Petal} \\ \hline
    $\boldsymbol{d_0}$ & \textbf{0.0021} & {0.0175}& {0.0062} & {0.0030} \\ \hline
    $\boldsymbol{d_1}$ & \textbf{0.0024} & {0.892} & {0.1116} & {0.0121}\\ \hline
    $\boldsymbol{d_2}$ &  \textbf{0.0040} & {3.405} & {0.5973} & {0.0914}\\  \hline
    $\boldsymbol{d_3}$ & \textbf{0.0034} & {0.9810} & {0.1578} & {0.0669} \\ \hline
    \multicolumn{5}{|c|}{Helix} \\ \hline
    $\boldsymbol{d_0}$ & {0.0043} & {0.1164} & {0.0032} & \textbf{0.0031} \\ \hline
    $\boldsymbol{d_1}$ & \textbf{0.0048} & {2.4147} & {0.1792} & {0.0162} \\ \hline
    $\boldsymbol{d_1}$&  \textbf{0.0079} & {6.9205} & {0.7451} & {0.1090}\\  \hline
    $\boldsymbol{d_2}$& \textbf{0.0038} & {1.568} & {0.2736} & {0.0682}\\ \hline

    %     $\boldsymbol{d_0}$ & \textbf{0.0885} & 0.4361 & 0.1265 & 0.0949\\ \hline
    % $\boldsymbol{d_1}$ & \textbf{0.0917} & 11.0973 & 3.5788 & 0.3700 \\ \hline
    % $\boldsymbol{d_2}$ & \textbf{0.1423} & 41.7936 & 18.1346 & 2.3365\\  \hline
    % $\boldsymbol{d_3}$ & \textbf{0.1075} & 14.4640 & 5.8893 & 2.4144 \\ \hline
    % \multicolumn{5}{|c|}{Petal} \\ \hline
    % $\boldsymbol{d_0}$ & \textbf{0.0663} & 0.5534 & 0.1960 & 0.0949 \\ \hline
    % $\boldsymbol{d_1}$ & \textbf{0.0759} & 28.2001 & 3.5283 & 0.3825\\ \hline
    % $\boldsymbol{d_2}$ & \textbf{0.1265} & 107.6465 & 18.8580 & 2.8867\\  \hline
    % $\boldsymbol{d_3}$ & \textbf{0.1075} & 31.0041 & 4.9881 & 2.1167 \\ \hline
    % \multicolumn{5}{|c|}{Helix} \\ \hline
    % $\boldsymbol{d_0}$ & 0.1358 & 3.6781 & 0.1012 & \textbf{0.0980} \\ \hline
    % $\boldsymbol{d_1}$ & \textbf{0.1517} & 76.3573 & 5.6594 & 0.5118 \\ \hline
    % $\boldsymbol{d_1}$ & \textbf{0.2497} & 218.6941 & 23.5399 & 3.4500 \\  \hline
    % $\boldsymbol{d_2}$ & \textbf{0.1202} & 49.5374 & 8.6483 & 2.1556 \\ \hline
    \end{tabular}

\end{table}

% rewrite this 
In summary, the proposed momentum-based Koopman framework achieves superior prediction accuracy and closed-loop tracking performance compared to the state-of-the-art alternatives. Further, the integration of GESOs enables the proposed architecture to maintain robust trajectory tracking even under substantial external disturbances.

\subsection{Experimental Results}\label{sec:exp}
This subsection evaluates the closed-loop prediction performance of the proposed scheme on FRANKA Research 3 manipulator arm. For learning the Koopman model, 500 trajectories are generated randomly within the workspace of the robot, each consisting of 1000 data points sampled at 200 Hz. The parameters for GESO \eqref{eq:geso} are $k_1 = 100$ and $k_2 = 2000$. The effect of the proposed algorithms is evaluated by implementing the control architecture to trace a hypotrochoid and a tricuspid path.
%The corresponding results are shown in Fig.~\ref{fig:franka}.

As seen from Fig.~\ref{fig:franka}, the robot is able to trace the desired path accurately both in the presence and absence of the external disturbances, while the NLK model is unable to trace the desired paths in either case. The resulting RMSE errors for the proposed scheme for the hypotrochoid shape are $0.024 \;\text{m}$ and $0.035 \;\text{m}$ for the case with and without disturbance, respectively. The corresponding errors for NLK are $ 0.077\;\text{m}$ and $0.091 \;\text{m}$. Similarly for the tricuspid shape, the resulting RMSE errors for the proposed scheme are $0.011 \;\text{m}$ and $0.019 \;\text{m}$ for the case with and without disturbance, respectively. The corresponding errors for NLK are $ 0.043\;\text{m}$ and $0.048 \;\text{m}$. The poor tracking performance of the NLK stems from a combination of the poor prediction model and a lack of effective disturbance compensation. Hence, the experimental results demonstrate accurate real-time trajectory tracking of the proposed scheme with the added benefit of mitigating disturbances. 

\section{Conclusion}\label{sec:conclusion}
This paper presents a momentum-based Koopman operator formulation for Euler-Lagrangian dynamics that employs an implicit generalized momentum–based state space. This representation decouples a known linear actuation channel from the state-dependent dynamics, making the dynamics more amenable to linear Koopman modeling in contrast to explicit state space representations, which lend themselves more naturally to bilinear Koopman formulations. By leveraging this structural separation, the proposed approach requires to learns only the unactuated dynamics rather than the full actuation-dependent system, thereby significantly reducing the number of learnable parameters, improving data efficiency, and lowering overall model complexity. To realize this framework, we introduced two neural network architectures capable of constructing Koopman embeddings from either actuated or unactuated data, enabling flexible deployment under diverse data availability scenarios. Linear Generalized Extended State Observer (GESO) is integrated with proposed scheme to enhance robustness to external disturbances and unmodeled dynamics. Unlike higher-order nonlinear observers employed in previous studies, linear GESOs can be seamlessly incorporated into linear control frameworks, preserving their structure while enhancing stability and robustness. Extensive simulation and experimental studies are conducted to show that the proposed formulation achieves superior predictive accuracy compared to explicit state space representation-based linear and bilinear Koopman models, while the integrated GESO ensures reliable performance under a wide range of disturbances.

\bibliographystyle{ieeetr}
\bibliography{citation}

\end{document}